\documentclass[runningheads]{llncs}

\usepackage{graphicx}
\usepackage{amsmath}
\usepackage{booktabs}
\usepackage{multirow}
\usepackage{rotating}
\usepackage{xcolor}
\usepackage{textcomp}
\usepackage{caption}
\usepackage{wrapfig}
\usepackage{xspace}

\makeatletter
\DeclareRobustCommand\onedot{\futurelet\@let@token\@onedot}
\def\@onedot{\ifx\@let@token.\else.\null\fi\xspace}

\def\eg{\emph{e.g}\onedot} 
\def\ie{\emph{i.e}\onedot} 
 
\def\etc{\emph{etc}\onedot} 
 
\def\etal{\emph{et al}\onedot}
\makeatother

\usepackage{pifont}
\newcommand{\cmark}{\ding{51}}%
\newcommand{\xmark}{\ding{55}}%

\usepackage[pagebackref,breaklinks,colorlinks]{hyperref}

\usepackage[capitalize]{cleveref}
\crefname{section}{Sec.}{Secs.}
\Crefname{section}{Section}{Sections}
\Crefname{table}{Table}{Tables}
\crefname{table}{Tab.}{Tabs.}

\usepackage{enumitem}
\setlist{nolistsep}

\graphicspath{{figures/}}

\usepackage{graphicx}
\usepackage{tikz}
\usepackage{comment}
\usepackage{amsmath,amssymb} 
\usepackage{color}

\begin{document}
\pagestyle{headings}
\mainmatter

\title{LESS: Label-Efficient Semantic Segmentation for LiDAR Point Clouds} 

\titlerunning{LESS: Label-Efficient Semantic Segmentation for LiDAR Point Clouds}

\author{Minghua Liu\inst{1}\thanks{Work done during internship at Waymo LLC.} \and
Yin Zhou \inst{2}\thanks{Corresponding to \email{yinzhou@waymo.com}.} \and
Charles R. Qi \inst{2} \and
Boqing Gong \inst{3} \and
Hao Su \inst{1} \and
Dragomir Anguelov \inst{2}}
\authorrunning{M. Liu et al.}

\institute{$^{1}$UC San Diego, $^{2}$Waymo, $^{3}$Google} 

\maketitle

\begin{abstract}
\vspace{-1em}
Semantic segmentation of LiDAR point clouds is an important task in autonomous driving. However, training deep models via conventional supervised methods requires large datasets which are costly to label. It is critical to have label-efficient segmentation approaches to scale up the model to new operational domains or to improve performance on rare cases. While most prior works focus on indoor scenes, we are one of the first to propose a label-efficient semantic segmentation pipeline for outdoor scenes with LiDAR point clouds. Our method co-designs an efficient labeling process with semi/weakly supervised learning and is applicable to nearly any 3D semantic segmentation backbones. Specifically, we leverage geometry patterns in outdoor scenes to have a heuristic pre-segmentation to reduce the manual labeling and jointly design the learning targets with the labeling process. In the learning step, we leverage prototype learning to get more descriptive point embeddings and use multi-scan distillation to exploit richer semantics from temporally aggregated point clouds to boost the performance of single-scan models. Evaluated on the SemanticKITTI and the nuScenes datasets, we show that our proposed method outperforms existing label-efficient methods. With extremely limited human annotations (\eg, 0.1\% point labels), our proposed method is even highly competitive compared to the fully supervised counterpart with 100\% labels.

\end{abstract}


\vspace{-2em}
\section{Introduction}
\vspace{-0.8em}

Light detection and ranging (LiDAR) sensors have become a necessity for most autonomous vehicles. They capture more precise depth measurements and are more robust against various lighting conditions compared to visual cameras. Semantic segmentation for LiDAR point clouds is an indispensable technology as it provides fine-grained scene understanding, complementary to object detection. For example, semantic segmentation help self-driving cars distinguish drivable and non-drivable road surfaces and reason about their functionalities, like parking areas and sidewalks, which is beyond the scope of modern object detectors.
 
\begin{figure}[t]
  \centering
   \includegraphics[width=0.8\linewidth]{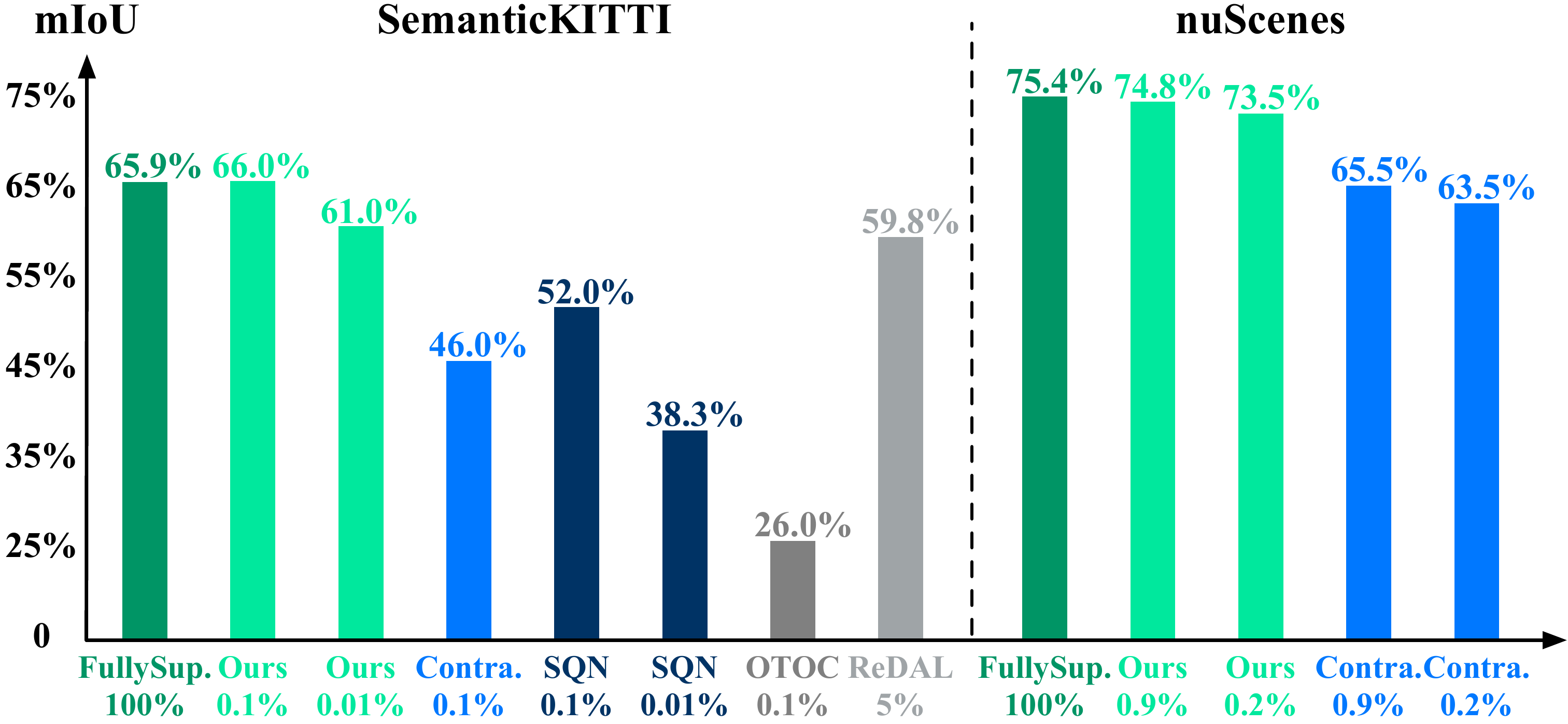}
   \caption{We compare LESS with Cylinder3D~\cite{zhu2021cylindrical} (our fully-supervised counterpart), ContrastiveSceneContext~\cite{hou2021exploring}, SQN~\cite{hu2021sqn}, OneThingOneClick~\cite{liu2021one}, and ReDAL~\cite{wu2021redal} on the SemanticKITTI~\cite{behley2019semantickitti} and nuScenes~\cite{caesar2020nuscenes} validation sets. The ratio between labels used and all points is listed below each bar. Please note that all competing label-efficient methods mainly focus on indoor settings and are not specially designed for outdoor LiDAR segmentation.}
   \label{fig:teaser}
\end{figure}

Based on large-scale public driving-scene datasets~\cite{behley2019semantickitti,caesar2020nuscenes}, several LiDAR semantic segmentation approaches have recently been developed~\cite{zhu2021cylindrical,xu2021rpvnet,cheng20212,yan2020sparse,tang2020searching}. Typically, these methods require fully labeled point clouds during training. Since a LiDAR sensor may perceive millions of points per second, exhaustively labeling all points is extremely laborious and time-consuming. Moreover, it may fail to scale when we extend the operational domain (\eg, various cities and weather conditions) and seek to cover more rare cases. Therefore, to scale up the system, it is critical to have label-efficient approaches for LiDAR semantic segmentation, whose goal is to minimize the quantity of human annotations while still achieving high performance.

While there are some prior works studying label-efficient semantic segmentation, they mostly focus on indoor scenes~\cite{dai2017scannet,armeni2017joint} or 3D object parts~\cite{chang2015shapenet}, which are quite different in point cloud appearance and object type distribution, compared to the outdoor driving scenes (\eg, significant variances in point density, extremely unbalanced point counts between common types, like ground and vehicles, and less common ones, such as cyclists and pedestrians). Besides, most prior explorations tend to address the problem from two independent perspectives, which may be less effective in our outdoor setting. Specifically, one perspective is improving labeling efficiency, where the methods resort to active learning~\cite{shi2021label,wu2021redal,luo2018semantic}, weak labels~\cite{ren20213d,wei2020multi}, and 2D supervision~\cite{wang2020weakly} to reduce labeling efforts. The other perspective focuses on training, where the efforts assume the partial labels are given and design semi/weakly supervised learning algorithms to exploit the limited labels and strive for better performance~\cite{liu2021one,xu2019semantic,ren20213d,xu2020weakly,guinard2017weakly,luo2018semantic,zhao2021few}. 

This paper proposes a novel framework, label-efficient semantic segmentation (LESS), for LiDAR point clouds captured by self-driving cars. Different from prior works, our method co-designs the labeling process and the model learning. Our co-design is based on two principles: 1) the labeling step is designed to provide bare minimum supervision, which is suitable for state-of-the-art semi/weakly supervised segmentation methods; 2) the model training step can tap into the labeling policy as a prior and deduce more learning targets. The proposed method can fit in a straightforward way with most state-of-the-art LiDAR segmentation backbones without introducing any network architectural change or extra computational complexity when deployed onboard. Our approach is suitable for effectively labeling and learning from scratch. It is also highly compatible with mining long-tail instances, where, in practice, we mainly want to identify and annotate rare cases based on trained models.

Specifically, we leverage a philosophy that outdoor-scene objects are often well-separated when isolating ground points and design a heuristic approach to pre-segment an outdoor scene into a set of connected components. The component proposals are of high purity (\ie, only contain one or a few classes) and cover most of the points. Then, instead of meticulously labeling all points, the annotators are only required to label one point per class for each component. In the model learning process, we train the backbone segmentation network with the sparse labels directly annotated by humans as well as the derived labels based on component proposals. To encourage a more descriptive embedding space, we employ contrastive prototype learning~\cite{gao2021contrastive,li2020prototypical,snell2017prototypical,yang2018robust,liu2021one}, which increases intra-class similarity and inter-class separation. We also leverage a multi-scan teacher model to exploit richer semantics within the temporally fused point clouds and distill the knowledge to boost the performance of the single-scan model.

We evaluate the proposed method on two large-scale autonomous driving datasets, SemanticKITTI~\cite{behley2019semantickitti} and nuScenes~\cite{caesar2020nuscenes}. We show that our method significantly outperforms existing label-efficient methods (see~\cref{fig:teaser}). With extremely limited human annotations, such as 0.1\% labeled points, the approach achieves highly competitive performance compared to the fully supervised counterpart, demonstrating the potential of practical deployment.

In summary, our contribution mainly includes:  \begin{itemize}[leftmargin=*]
    \item Analyze how label-efficient segmentation of outdoor LiDAR point clouds differs from the indoor settings, and show that the unbalanced category distribution is one of the main challenges.
    \item Leverage the unique geometric structure of LiDAR point clouds and design a heuristic algorithm to pre-segment input points into high-purity connected components. A customized labeling policy is then proposed to exploit the components with tailored labels and losses. 
   \item Adapt beneficial components into label-efficient LiDAR segmentation and carefully design a network-agnostic pipeline that achieves on-par performance with the fully supervised counterpart.
    \item Evaluate the proposed pipeline on two large-scale autonomous driving datasets and extensively ablate each module.
\end{itemize}


\vspace{-1em}
\section{Related work}
\vspace{-0.8em}

\subsection{Segmentation networks for LiDAR point clouds}
\vspace{-0.5em}

In contrast to indoor-scene point clouds, outdoor LiDAR point clouds' large scale, varying density, and sparsity require the segmentation networks to be more efficient. Many works project the 3D point clouds from spherical view ~\cite{razani2021lite,kochanov2020kprnet,cortinhal2020salsanext,duerr2020lidar,xu2020squeezesegv3,alonso20203d,li2020multi,milioto2019rangenet++}(\ie, range images) or bird's-eye-view~\cite{rist2020scssnet,zhang2020polarnet} onto 2D images, or try to fuse different views~\cite{liong2020amvnet,gerdzhev2021tornado,alnaggar2021multi}. There are also some works directly consuming point clouds~\cite{thomas2019kpconv,fang2020spatial,hu2020randla,cheng2020cascaded}. They aim to structure the irregular data more efficiently. Zhu~\etal~\cite{zhu2021cylindrical} employ the voxel-based representation and alleviate the high computational burden by leveraging cylindrical partition and sparse asymmetrical convolution. Recent works also try to fuse the point and voxel representations~\cite{tang2020searching,yan2020sparse,zhang2020deep,cheng20212}, and even with range images~\cite{xu2021rpvnet}. All of these works can serve as the backbone network in our label-efficient framework.

\vspace{-1em}
\subsection{Label-efficient 3D semantic segmentation}
\vspace{-0.5em}
Label-efficient 3D semantic segmentation has recently received lots of attention~\cite{gao2020we}. Previous explorations are mainly two-fold: labeling and training.

As for labeling, several approaches seek active learning~\cite{shi2021label,wu2021redal,luo2018semantic},  which iteratively selects and requests points to be labeled during the network training. Hou \etal~\cite{hou2021exploring} utilize features from unsupervised pre-training to choose points for labeling. Wang~\etal~\cite{wang2020weakly} project the point clouds to 2D and leverage 2D supervision signals. Some works utilize scene-level or sub-cloud-level weak labels~\cite{ren20213d,wei2020multi}. There are also several approaches using rule-based heuristics or handcrafted features to help annotation~\cite{mei2019incorporating,thomas2021self,guinard2017weakly}.

As for training, Xie \etal~\cite{hou2021exploring,xie2020pointcontrast} utilize contrastive learning for unsupervised pre-training. Some approaches employ self-training to generate pseudo-labels~\cite{liu2021one,xu2019semantic,ren20213d}. Lots of works use Conditional Random Fields (CRFs)~\cite{liu2021one,xu2020weakly,guinard2017weakly,luo2018semantic} or random walk~\cite{zhao2021few} to propagate labels. Moreover, there are also works that utilize prototype learning~\cite{liu2021one,zhao2021few}, siamese learning~\cite{xu2020weakly,ren20213d}, temporal constraints~\cite{mei2019semantic}, smoothness constraints~\cite{ren20213d,shi2021label},  attention~\cite{wei2020multi,zhao2021few}, cross task consistency~\cite{ren20213d}, and synthetic data~\cite{xiao2021synlidar} to help training.

However, most recent works mainly focus on indoor scenes~\cite{dai2017scannet,armeni2017joint} or 3D object parts~\cite{chang2015shapenet}, while outdoor scenarios are largely under-explored.

\setlength{\belowdisplayskip}{3pt} \setlength{\belowdisplayshortskip}{3pt}
\setlength{\abovedisplayskip}{3pt} \setlength{\abovedisplayshortskip}{3pt}

\vspace{-0.6em}
\section{Method}
\vspace{-0.6em}

In this section, we present our LESS framework. Since existing label-efficient segmentation works typically address domains other than autonomous driving, we first conduct a pilot study to understand the challenges in this novel setting and introduce motivations behind LESS (\cref{sec:pilot_study}). After briefly going over our LESS framework (\cref{sec:overview}), we dive into the details of each part (\cref{sec:pre-segmentation,sec:various-labels,sec:prototype-learning,sec:multi-scan-distillation}).

\vspace{-1em}
\subsection{Pilot study: what should we pay attention to?}
\vspace{-0.5em}
\label{sec:pilot_study}

\begin{table}[t]
  \centering
  \scriptsize
    \begin{tabular}{c|ccccccc}
    \hline
    dataset & vegetation & road  & building & car   & motorcycle & person & bicycle \\
    \hline
    SemanticKITTI & 1606  & 1197  & 799   & 257   & 2     & 2     & 1 \\
    nuScenes & 867 & 2242 & 1261 & 270 & 3 & 16 & 1 \\
    \hline
    \end{tabular}%
\caption{\textbf{Point distribution across the most common and rarest categories of SemanticKITTI~\cite{behley2019semantickitti} and nuScenes~\cite{caesar2020nuscenes}.} Numbers are normalized by the sample quantity of bicycles.}
  \label{tab:kitti_distribution}%
  \vspace{-2em}
\end{table}%

Previous works~\cite{liu2021one,hou2021exploring,shi2021label,wang2020weakly,ren20213d,wei2020multi,xu2020weakly,zhao2021few} on label-efficient 3D semantic segmentation mainly focused on indoor datasets, such as ScanNet-v2~\cite{dai2017scannet} and S3DIS~\cite{armeni2017joint}. In these datasets, input points are sampled from high-quality reconstructed meshes and are thus densely and uniformly distributed. Also, objects in indoor scenarios typically share similar sizes and have a relatively balanced class distribution. However, in outdoor settings, input point clouds demonstrate substantially higher complexity due to the varying point density and the ubiquitous occlusions throughout the scene. Moreover, in outdoor driving scenes, the sample distribution across different categories is highly unbalanced due to factors including occurring frequency and object size. \cref{tab:kitti_distribution} shows the point distribution over two autonomous driving datasets, where the numbers of road points are 1,197 and 2,242 times larger than that of bicycle points, respectively. The extremely unbalanced distribution adds extra difficulty for label-efficient segmentation, whose goal is to only label a tiny portion of points.

\begin{wrapfigure}{r}{0.5\textwidth}
  \centering
  \vspace{-2em}
   \includegraphics[width=\linewidth]{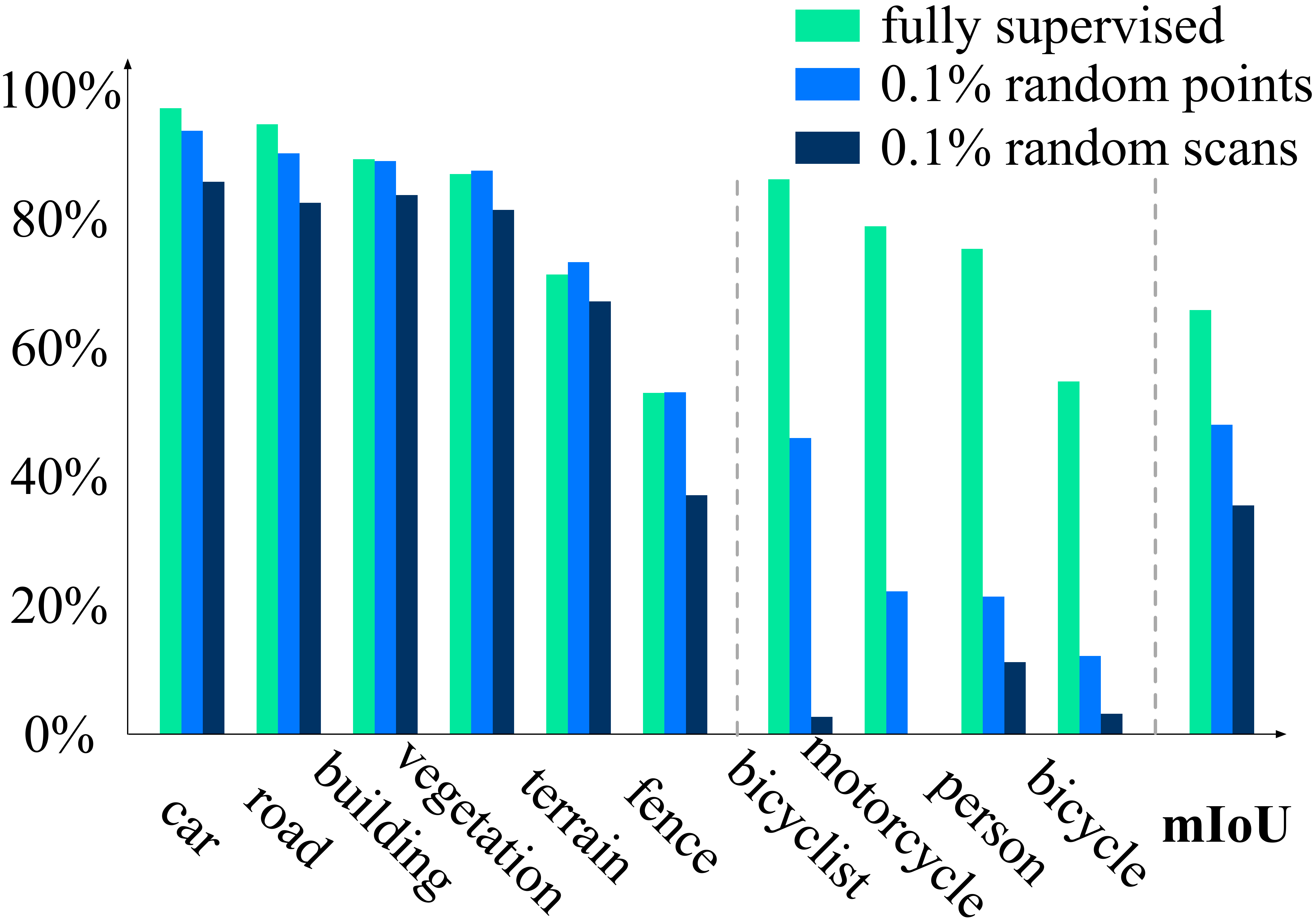}
   \caption{\textbf{Pilot study: performances (IoU) of the most common and rarest categories.}  Models are trained with 100\% of labels and 0.1\% of labels (in terms of points or scans) on SemanticKITTI~\cite{behley2019semantickitti}.}
   \vspace{-2em}
   \label{fig:pilot_study}
\end{wrapfigure}

We conduct a pilot study to further examine this challenge. Specifically, we train a state-of-the-art semantic segmentation network, Cylinder3D~\cite{zhu2021cylindrical}, on the  SemanticKITTI dataset with three intuitive setups: (a) $100\%$ labels, (b) randomly annotating 0.1\% points per scan, and (c) randomly selecting 0.1\% scans and annotating all points for the selected scans. The results are shown in~\cref{fig:pilot_study}. Without any special efforts, ``0.1\% random points'' can already achieve a mean IoU of 48.0\%, compared to 65.9\% by the fully supervised version. On common categories, such as car, road, building, and vegetation, the performances of the ``0.1\% label'' models are close to the fully supervised model. However, on the underrepresented categories, such as bicycle, person, and motorcycle, we observe substantial performance gaps compared to the fully supervised model. These categories tend to have small sizes, appear less frequently, and are thus more vulnerable when reducing the annotation budget. However, they are still critical for many applications such as autonomous driving. Moreover, we find that ``0.1\% random points'' outperforms ``0.1\% random scans'' by a large margin, mainly due to its label diversity.

These observations inspire us to rethink the existing paradigm of label-efficient segmentation. While prior works typically focus on either efficient labeling or improving training approaches, we argue that it can be more effective to address the problem by co-designing both. By integrating the two parts, we may cover more underrepresented instances with a limited labeling budget, and exploit the labeling efforts more effectively during network training.

\vspace{-1.3em}
\subsection{Overview}
\vspace{-0.5em}

\label{sec:overview}

\begin{figure*}[t]
  \centering
   \includegraphics[width=\linewidth]{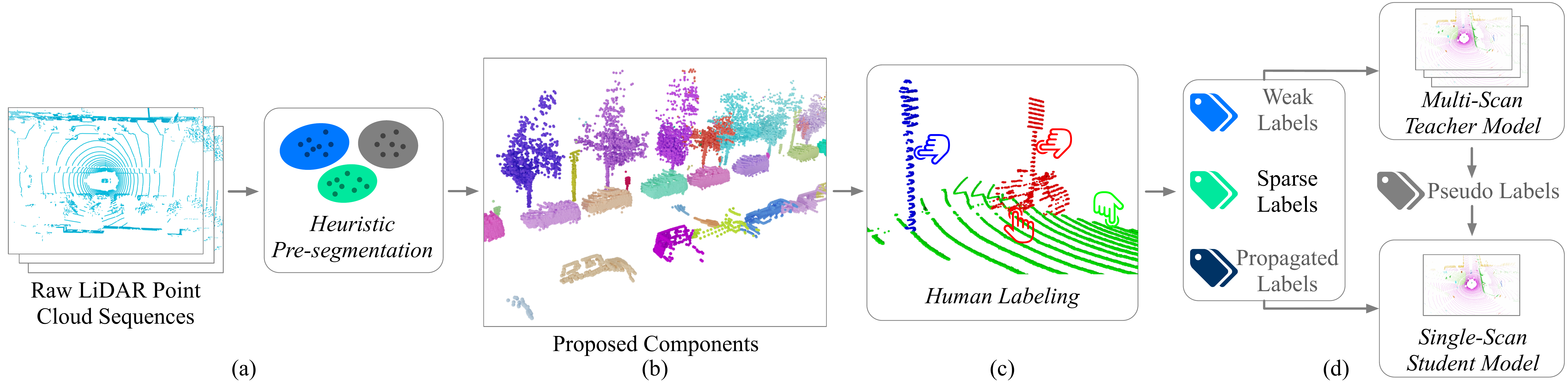}
   \caption{\textbf{Overview of our LESS pipeline.} (a) We first utilize a heuristic algorithm to pre-segment each LiDAR sequence into a set of connected components. (b) Examples of the proposed components. Different colors indicate different components. For clear visualization, components of ground points are not shown. (c) Human annotators only need to coarsely label each component. Each color denotes a proposed component, and each click icon indicates a labeled point. Only sparse labels are directly annotated by humans. (d) We then train the network to digest various labels and utilize multi-scan distillation to exploit richer semantics in the temporally fused point clouds.}
   \label{fig:overview}
\end{figure*}

Our LESS framework integrates pre-segmentation, labeling, and network training. It can work with most existing LiDAR segmentation backbones without changing their network architectures or inference latency.  As shown in~\cref{fig:overview}, our pipeline takes raw LiDAR sequences as input. It first employs a heuristic method to partition the point clouds into a set of high-purity components (\cref{sec:pre-segmentation}). Instead of exhaustively labeling all points, annotators only need to quickly label a few points for each component proposal (\eg, one point label for each class that appears). Besides the human-annotated sparse labels, we derive other types of labels so as to train the network with more context information (\cref{sec:various-labels}).  During the network training, we employ contrastive prototype learning to realize a more descriptive embedding space (\cref{sec:prototype-learning}). We also boost the single-scan model by distilling the knowledge from a multi-scan teacher, which exploits richer semantics within the temporally fused point clouds (\cref{sec:multi-scan-distillation}).

\vspace{-1.3em}
\subsection{Pre-segmentation}
\vspace{-0.8em}

\label{sec:pre-segmentation}

We design a heuristic pre-segmentation to subdivide the point cloud into a collection of components. Each resulting component proposal is of high purity, containing only one or a few categories, which facilitates annotators to coarsely label all the proposals, \ie, one point label per class (\cref{sec:various-labels}). In this way, we can derive dense supervision by disseminating the sparse point-wise annotations to the whole components. Since modern networks can learn the semantics of homogeneous neighborhoods from sparse annotations, spending lots of annotation budgets on large objects may be futile. Our component-wise coarse annotation is agnostic to the object size, which benefits underrepresented small objects. 

For indoor scenarios, many prior arts ~\cite{liu2021one,guinard2017weakly,shi2021label} leverage the surface normal and color information to generate super voxels and assume that the points within each super voxel share the same category. These approaches, however, might not generalize to outdoor LiDAR point clouds, where the surface can be noisy and color information is not available. Since the homogeneity assumption is hard to hold, we instead propose to lift this constraint and allow each component to contain more than one category. 

Unlike indoor scenarios, objects in outdoor scans are often well-separated after detecting and isolating the ground points. Inspired by this philosophy, we design an intuitive approach to pre-segment each LiDAR sequence, which includes four steps: \textbf{(a) Fuse overlapping scans.} We first split a LiDAR sequence into sub-sequences, each containing $t$ consecutive scans. We then fuse the scans of each sub-sequence based on the provided ego-poses. In this way, we can label the same instance across overlapping scans at one click. \textbf{(b) Detect ground points.} While the ground surface may not be flat at the full-scene scale, we assume for each local region (\eg, 5m $\times$ 5m), the ground points can be fitted by a plane. We thus partition the whole scene into a uniform grid according to the $xy$ coordinates, and then employ the RANSAC algorithm~\cite{fischler1981random} to detect the ground points for each local cell. Since the ground points may belong to different categories (\eg, parking zone, sidewalk, and road), we regard the ground points from each local cell as a single component instead of merging all of them. We allow a single ground component to contain multiple classes, and one point per class will be labeled later. \textbf{(c) Construct connected components.} After detecting and isolating the ground points, the remaining objects are often well-separated. We build a graph $G$, where each node represents a point. We connect every pair of points $(u, v)$ in the graph, whose Euclidean distance is smaller than a threshold $\tau$. We then divide the points into groups by calculating the connected components for the graph $G$. Due to the non-uniform point density distribution of the LiDAR point clouds, it is hard to use a fixed threshold across different ranges. We thus propose an adaptive threshold $\tau(u,v) = \max(r_u, r_v)\times d$ to compensate for the varying density, where $r_u$ and $r_v$ are the distances between the points and the sensor centers, and $d$ is a pre-defined hyper-parameter. \textbf {(d) Subdivide large components.} After step (c), there usually exist some connected components covering an enormous area (\eg, buildings and vegetation), which are prone to include some small objects. To keep each component of high purity and facilitate network training, we subdivide oversized components to ensure each component is bounded within a fixed size. Also, we ignore small components with only a few points, which tend to be noisy and can lead to excessive component proposals.

In practice, we find our pre-segmentation generates a small number of components for each sequence. The component proposals cover most of the points, and each component tends to have high purity. These open up the possibility of quickly bootstrapping the labeling from scratch. Moreover, unlike other methods~\cite{liu2021one,guinard2017weakly,shi2021label} relying on various handcrafted features, our method only utilizes the simple geometrical connectivity, allowing it to generalize to various scenarios without tuning lots of hyper-parameters. Please refer to ~\cref{sec:exp-pre-seg} for statistics of the pre-segmentation results and the supplementary material for more details. 

\vspace{-1.3em}
\subsection{Annotation policy \& training labels}
\vspace{-0.8em}

\label{sec:various-labels}

Instead of meticulously labeling every point, we propose to coarsely annotate the component proposals. Specifically, for each component proposal, an annotator needs to first skim through the component and then label only one point for each identified category. \cref{fig:overview} (c) illustrates an example where the pre-segmentation yields three components colored in red, blue, and green, respectively. Because the blue component only has traffic-sign points, the annotator only needs to randomly select one point to label. The green component is similar, as it only contains road points. In the red component, there is a bicycle lying against a traffic sign, and the annotator needs to select one point for each class to label. By coarsely labeling all components, we are unlikely to miss any underrepresented instances, as the proposed components cover the majority of points. Moreover, since the number of components is orders of magnitude smaller than that of points and our coarse annotation policy frees annotators from carefully labeling instance boundaries (required in the labeling process to build SemanticKITTI~\cite{behley2019semantickitti} dataset), we are thus able to reduce manual labeling costs. 

Based on the component proposals, we can obtain three types of labels. \textbf{Sparse labels:} points directly labeled by annotators. Although only a tiny subset of points are labeled, sparse labels provide the most accurate and diverse supervision. \textbf{Weak labels:} classes that appear in each component. Weak labels are derived based on human-annotated sparse labels within each component. In the example of \cref{fig:overview} (c), all red points can only be either bicycles or traffic signs. We disseminate weak labels from each component to the points therein. The multi-category weak labels provide weak but dense supervision and cover most points. \textbf{Propagated labels:} for the pure components (\ie, only one category appears), we can propagate the label to the entire component. Given the effectiveness of our pre-segmentation approach, the propagated labels also cover a wide range of points. However, since some categories may be easier to be separated and prone to form pure components, the distribution of the propagated labels may be biased and less diverse than the sparse labels. 

We formulate a joint loss function by exploiting the three types of labels: 
$
\mathcal{L} = \mathcal{L}_{\text{sparse}} + \mathcal{L}_{\text{propagated}} + \mathcal{L}_{\text{weak}} 
\label{equ:loss}
$, where $\mathcal{L}_{\text{sparse}}$ and $\mathcal{L}_{\text{propagated}}$ are weighted cross-entropy loss with respect to the sparse labels and propagated labels, respectively. We utilize inverse square root of label frequency ~\cite{mikolov2013distributed,zou2021towards,mahajan2018exploring} as category weights to emphasize underrepresented categories. Here, we calculate a cross-entropy loss for each label type separately,  because propagated labels significantly outnumber sparse labels while sparse labels provide more diverse supervision.

Denote the weak labels as binary masks $l_{ij}$ for point $i$ and category $j$. $l_{ij} = 1$ when point $i$ belongs to a component that contains category $j$. We exploit the multi-category weak labels by penalizing the impossible predictions:
\begin{equation}
    \mathcal{L}_{\text{weak}} = - \frac{1}{n}\sum_{i=1}^{n} \log(1-\sum_{l_{ij} = 0}p_{ij})
\end{equation}
where $p_{ij}$ is the predicted probability of point $i$, and $n$ is the number of points. Prior approaches~\cite{wei2020multi,ren20213d} aggregate per-point predictions into component-level predictions and then utilize the multiple-instance learning loss (MIL)~\cite{pathak2014fully,pinheiro2015image} to supervise the learning. Here, we only penalize the negative predictions without encouraging the positive ones. This is because our network takes a single-scan point cloud as input, but the labels are collected and derived over the temporally fused point clouds. Hence, a positive instance may not always appear in each individual scan, due to occlusions or limited sensor coverage. 

\vspace{-1em}
\subsection{Contrastive prototype learning}
\vspace{-0.8em}
\label{sec:prototype-learning}

Besides the great success in self-supervised representation learning~\cite{chen2020simple,he2020momentum,oord2018representation}, contrastive learning has also shown effectiveness in supervised learning and few-shot learning~\cite{khosla2020supervised,gao2021contrastive,schroff2015facenet,sohn2016improved}. It can overcome shortcomings of the cross-entropy loss, such as poor margins~\cite{elsayed2018large,liu2016large,khosla2020supervised,yang2018robust}, and construct a more descriptive embedding space.
Following~\cite{gao2021contrastive,li2020prototypical,snell2017prototypical,yang2018robust,liu2021one}, we exploit the limited annotations by learning distinctive class prototypes (\ie, class centroids in the feature space). Without pre-training, a contrastive prototype loss $\mathcal{L}_{\text{proto}}$ is added to ~\cref{equ:loss} as an auxiliary loss. Due to the limited annotations and unbalanced label distribution, only using samples within each batch to determine class prototypes may lead to unstable results. Inspired by the idea of momentum contrast~\cite{he2020momentum}, we instead learn the class prototypes $\mathbf{P}_{c}$ by using a moving average over iterations:
\begin{equation}
\mathbf{P}_{c} \leftarrow m \mathbf{P}_{c} + (1 - m) \frac{1}{n_c}\sum_{y_i = c} \operatorname{stopgrad}(h(f(x_i)))  
\end{equation}
where $f(x_i)$ is the embedding of point $x_i$, $h$ is a linear projection head with vector normalization, $\operatorname{stopgrad}$ denotes the stop gradient operation, $y_i$ is the label of $x_i$, $n_c$ is the number of points with label $c$ in a batch, and $m$ is a momentum coefficient. In the beginning, $\mathbf{P}_{c}$ are initialized randomly.  

The prototype loss $\mathcal{L}_{\text{proto}}$ is calculated for the points with sparse labels and propagated labels within each batch:
\begin{equation}
\mathcal{L}_{\text{proto}} = \frac{1}{n} \sum_i^n -w_{y_i} \log \frac{\exp(h(f(x_i)) \cdot \mathbf{P}_{y_i} / \tau)}{\sum_c \exp(h(f(x_i)) \cdot \mathbf{P}_{c} / \tau)}
\end{equation}
where $h(f(x_i)) \cdot \mathbf{P}_{y_i}$ indicates the cosine similarity between the projected embedding and the prototype, $\tau$ is a temperature hyper-parameter, $n$ is the number of points, and $w_{y_i}$ is the inverse square root weight of category $y_i$. $\mathcal{L}_{\text{proto}}$ aims to learn a better embedding space by increasing intra-class compactness and inter-class separability.

\vspace{-1.3em}
\subsection{Multi-scan distillation}
\vspace{-0.8em}

\label{sec:multi-scan-distillation}
We aim to learn a segmentation network that takes a single LiDAR scan as input and can be deployed in real-time onboard applications. During our label-efficient training, we can train a multi-scan network as a teacher model. It applies temporal fusion of multiple scans and takes the densified point cloud as input, compensating for the sparsity and incompleteness within a single scan. The teacher model is thus expected to exploit the richer semantics and perform better than a single-scan model. Especially, it may improve the performance for those underrepresented categories, which tend to be small and sparse. After that, we distill the knowledge from the multi-scan teacher model to boost the performance of the single-scan student model. 

Specifically, for a scan at time $t$, we fuse the point clouds of neighboring scans at time $\{t + i\Delta; i \in [-2, 2]\}$ ($\Delta$ is a time interval) using the ego-poses of the LiDAR sensor. To enable a large batch size, we use voxel subsampling~\cite{Zhou_2018_CVPR} to normalize the fused point cloud to a fixed size. Labels are then fused accordingly. Besides the spatial coordinates, we also concatenate an additional channel indicating the time index $i$ of each point. The teacher model is trained using the loss functions introduced in~\cref{sec:various-labels,sec:prototype-learning}.

The student model shares the same backbone network and is first trained from scratch in the same way as the teacher model except for the single-scan input. We then fine-tune it by incorporating an additional distillation loss $\mathcal{L}_{\text{dis}}$. Specifically, following~\cite{hinton2015distilling}, we match student predictions with the soft pseudo-labels generated by the teacher model via a cross-entropy loss:
\begin{equation}
  \footnotesize
   \mathcal{L}_{\text{dis}} =
   -\frac{T^2}{n}\sum_{i}^{n}\sum_{c} \frac{\exp (u_{ic} / T)}{\sum_{c^{\prime}} \exp(u_{ic^{\prime}} / T)} \log \left ( \frac{\exp (v_{ic} / T)}{\sum_{c^{\prime}} \exp(v_{ic^{\prime}} / T)} \right )
\end{equation}
where $u_{ic}$  and $v_{ic}$ are the predicted logits for point $i$ and category $c$ by the teacher and student models respectively, and $T$ is a temperature hyper-parameter. A higher temperature is typically used so that the probability distribution across classes is smoother, and the distillation is thus encouraged to match the negative logits, which also contain rich information. The cross-entropy is multiplied by $T^2$ to align the magnitudes of the gradients with existing other losses~\cite{hinton2015distilling}. 

Please note that the idea of multi-scan distillation may only be beneficial for our label-efficient LiDAR segmentation setting. For the fully supervised setting, all labels are already available and accurate, and there is no need to leverage the pseudo labels. For the indoor setting, all points are sampled from high-quality reconstructed meshes, and there is no need for a multi-scan teacher model.


\vspace{-1em}
\section{Experiments}
\vspace{-1em}

We employ Cylinder3D~\cite{zhu2021cylindrical}, a recent state-of-the-art method for LiDAR semantic segmentation, as our backbone network. We utilize ground truth labels to mimic the obtained human annotations, and no extra noise is added. Please refer to the supplementary material for more implementation and training details. 

We evaluate the proposed method on two large-scale autonomous driving datasets, SemanticKITTI~\cite{behley2019semantickitti} and nuScenes~\cite{caesar2020nuscenes}. \textbf{SemanticKITTI}~\cite{behley2019semantickitti} is collected in Germany with 64-beam LiDAR sensors. The (sensor) capture and annotation frequency is 10 Hz. It contains 10 training sequences (19k scans), 1 validation sequence (4k scans), and 11 testing sequences (20k scans). 19 classes are used for segmentation. \textbf{nuScenes}~\cite{caesar2020nuscenes} is collected in Boston and Singapore with 32-beam LiDAR sensors. Although the (sensor) capture frequency is 20Hz, the annotation frequency is only 2Hz. It contains 700 training sequences (28k scans), 150 validation sequences (6k scans), and 150 testing sequences (6k scans). 16 classes are used for segmentation. For both datasets, we follow the official guidance~\cite{behley2019semantickitti,caesar2020nuscenes} to use mean intersection-over-union (mIoU) as the evaluation metric.

\vspace{-1.3em}
\subsection{Comparison on SemanticKITTI}
\vspace{-0.8em}
\label{sec:exp-kitti}
We compare the proposed method with both label-efficient ~\cite{wu2021redal,hu2021sqn,liu2021one,hou2021exploring} and fully supervised~\cite{xu2020squeezesegv3,zhang2020polarnet,alnaggar2021multi,gan2020bayesian,duerr2020lidar,kochanov2020kprnet,tang2020searching,liong2020amvnet,zhu2021cylindrical} methods. Please note that all competing label-efficient methods mainly focus on indoor settings and are not specially designed for outdoor LiDAR segmentation. Among them, ContrastiveSC~\cite{hou2021exploring} employs contrastive learning as unsupervised pre-training and uses the learned features for active labeling, ReDAL~\cite{wu2021redal} also employs active labeling, OneThingOneClick~\cite{liu2021one} proposes a self-training approach and iteratively propagate the labels, and SQN~\cite{hu2021sqn} presents a network by leveraging the similarity between neighboring points. We report the results on the validation set. Since ContrastiveSC~\cite{hou2021exploring} and OneThingOneClick~\cite{liu2021one} are only tested on indoor datasets in the original paper, we adapt the source code published by the authors and train their models on SemanticKITTI~\cite{behley2019semantickitti}. For other methods, the results are either obtained from the literature or correspondences with the authors. 

\setlength\tabcolsep{1.6pt}
\begin{table*}[t]
  \centering
  \scriptsize
    \begin{tabular}{c|c|c|ccccccccccccccccccc}
    \toprule
    Method & Annot. & mIoU & \begin{sideways}car\end{sideways} & \begin{sideways}bicycle\end{sideways} & \begin{sideways}motorcycle\end{sideways} & \begin{sideways}truck\end{sideways} & \begin{sideways}other-vehicle\end{sideways} & \begin{sideways}person\end{sideways} & \begin{sideways}bicyclist\end{sideways} & \begin{sideways}motorcyclist\end{sideways} & \begin{sideways}road\end{sideways} & \begin{sideways}parking\end{sideways} & \begin{sideways}sidewalk\end{sideways} & \begin{sideways}other-ground\end{sideways} & \begin{sideways}building\end{sideways} & \begin{sideways}fence\end{sideways} & \begin{sideways}vegetation\end{sideways} & \begin{sideways}trunk\end{sideways} & \begin{sideways}terrain\end{sideways} & \begin{sideways}pole\end{sideways} & \begin{sideways}traffic-sign\end{sideways} \\
    \midrule
        SqueezeSegV3~\cite{xu2020squeezesegv3} & \multirow{9}[2]{*}{100\%} & 52.7  & 86  & 31  & 48  & 51  & 42  & 52  & 52  & 0   & 95  & 47  & 82  & 0   & 80  & 47  & 83  & 53  & 72  & 42  & 38 \\
    PolarNet~\cite{zhang2020polarnet} &       & 53.6  & 92  & 31  & 39  & 46  & 24  & 54  & 62  & 0   & 92  & 47  & 78  & 2   & 89  & 46  & 85  & 60  & 72  & 58  & 42 \\
    MPF~\cite{alnaggar2021multi}   &       & 57.0  & 94  & 28  & 55  & 62  & 36  & 57  & 74  & 0   & 95  & 47  & 81  & 1   & 88  & 53  & 86  & 54  & 73  & 57  & 42 \\
    S-BKI~\cite{gan2020bayesian} &       & 57.4  & 94  & 34  & 57  & 45  & 27  & 53  & 72  & 0   & 94  & 50  & 84  & 0   & 89  & 60  & 87  & 63  & 75  & 64  & 45 \\
    \scriptsize{TemporalLidarSeg}~\cite{duerr2020lidar} &       & 61.3  & 92  & 43  & 54  & 84  & 61  & 64  & 68  & 0  & 95  & 44  & 83  & 1   & 89  & 60  & 85  & 64  & 71  & 59  & 47 \\
    KPRNet~\cite{kochanov2020kprnet} &       & 63.1  & 95  & 43  & 60  & 76  & 51  & 75  & 81  & 0   & 96  & 51  & 84  & 0   & 90  & 60  & 88  & 66  & 76  & 63  & 43 \\
    SPVNAS~\cite{tang2020searching} &       & 64.7  & 97  & 35  & 72  & 81  & 66  & 71  & 86  & 0   & 94  & 48  & 81  & 0   & 92  & 67  & 88  & 65  & 74  & 64  & 49 \\
    AMVNet~\cite{liong2020amvnet} &       & 65.2  & 96  & 49  & 65  & 89  & 55  & 71  & 86  & 0   & 96  & 54  & 83  & 0   & 91  & 62  & 88  & 67  & 74  & 65  & 49 \\
    \textbf{Cylinder3D}~\cite{zhu2021cylindrical} &       & 65.9  & 97  & 55  & 79  & 80  & 67  & 75  & 86  & 1   & 95  & 46  & 82  & 1   & 89  & 53  & 87  & 71  & 71  & 66  & 53 \\
    \textbf{Cylinder3D$^{\star}$} & &  66.2  & 97  & 48  & 72  & 94  & 67  & 74  & 91  & 0   & 93  & 44  & 79  & 3   & 91  & 60  & 88  & 70  & 72  & 63  & 53 \\   
    \midrule
    ReDAL~\cite{wu2021redal} & 5\%   & 59.8  & 95  & 30  & 59  & 63  & 50  & 63  & 84  & 1   & 92  & 39  & 78  & 1   & 89  & 54  & 87  & 62  & 74  & 64  & 50 \\
 \scriptsize{OneThingOneClick}~\cite{liu2021one} & 0.1\% &  26.0  & 77  & 0   & 0   & 2   & 1   & 0   & 2   & 0   & 63  & 0   & 38  & 0   & 73  & 44  & 78  & 39  & 53  & 25  & 0 \\
  ContrastiveSC~\cite{hou2021exploring} & 0.1\% &       46.0  & 93  & 0   & 0   & 62  & 45  & 28  & 0   & 0   & 90  & 39  & 71  & 6   & 90  & 42  & 89  & 57  & 75  & 54  & 34 \\ 
    SQN~\cite{hu2021sqn}   & 0.1\% & 52.0  & 93  & 8   & 35  & 59  & 46  & 41  & 59  & 0   & 91  & 37  & 76  & 1   & 89  & 51  & 85  & 61  & 73  & 53  & 35 \\
    LESS (Ours)  & 0.1\% & \textbf{66.0}  & 97  & 50  & 73  & 94  & 67  & 76  & 92  & 0   & 93  & 40  & 79  & 3   & 91  & 60  & 87  & 68  & 71  & 62  & 51   \\
    SQN~\cite{hu2021sqn}   & 0.01\% & 38.3  & 83  & 0   & 22  & 12  & 17  & 15  & 47  & 0   & 85  & 21  & 65  & 0   & 79  & 37  & 77  & 46  & 67  & 44  & 12 \\
    LESS (Ours)  & 0.01\% & \textbf{61.0}  & 96  & 33  & 61  & 73  & 59  & 68  & 87  & 0   & 92  & 38  & 76  & 5   & 89  & 52  & 87  & 67  & 71  & 59  & 46 \\
    \bottomrule
    \end{tabular}%
\caption{\textbf{Comparison on the SemanticKITTI validation set.} Cylinder3D~\cite{zhu2021cylindrical} is our fully supervised counterpart. Cylinder3D$^{\star}$ is our re-trained version with our proposed prototype learning and multi-scan distillation.}
  \label{tab:semantic_kitti}%
  \vspace{-1em}
\end{table*}%

\begin{figure*}[t]
  \centering
   \includegraphics[width=\linewidth]{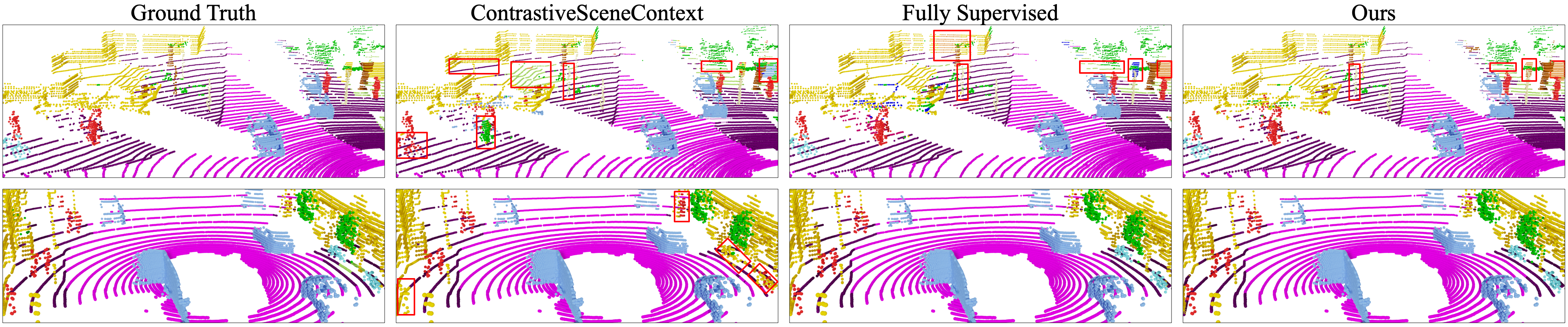}
   \caption{\textbf{Qualitative examples on the SemanticKITTI~\cite{behley2019semantickitti} (first row) and nuScenes~\cite{caesar2020nuscenes} (second row) validation sets.} Please zoom in for the details. Red rectangles highlight the wrong predictions. Our results are similar to the fully supervised counterpart, while ContrastiveSceneContext~\cite{hou2021exploring} produces worse results on underrepresented categories (see persons and bicycles). Please noet that, points in two datasets (with different density) are visualized in different point size for better visualization.}
   \label{fig:qualitative_examples}
\end{figure*}

\cref{tab:semantic_kitti} lists the results, where our method outperforms existing label-efficient methods by a large margin. With only 0.1\% sparse labels (as defined in~\cref{sec:various-labels}), it even completely match the performance of the fully supervised baseline Cylinder3D~\cite{zhu2021cylindrical}, which demonstrates the potential of deployment into real applications. By checking the breakdown results, we find that the differences between methods mainly come from the underrepresented categories, such as bicycle, motorcycle, person, and bicyclist. Existing label-efficient methods, which are mainly designed for indoor settings, suffer a lot from the highly unbalanced sample distribution, while our method is remarkably competitive in those underrepresented classes. See~\cref{fig:qualitative_examples} for further demonstration. OneThingOneClick~\cite{liu2021one} fails to produce decent results, which is partially due to its pure super-voxel assumption that does not always hold in outdoor scenes. As for the 0.01\% annotations setting, the performance of SQN~\cite{hu2021sqn} drops drastically to 38.3\%, whereas our proposed method can still achieve a high mIoU of 61.0\%. 
For completeness, we also re-train Cylinder3D~\cite{zhu2021cylindrical} with our proposed prototype learning and multi-scan distillation. We find that the two strategies provide marginal gain in the fully-supervised setting, where all labels are available and accurate.

\vspace{-1.3em}
\subsection{Comparison on nuScenes}
\vspace{-0.8em}
\label{sec:exp-nuscenes}

We also compare the proposed method with existing approaches on the nuScenes~\cite{caesar2020nuscenes} dataset and report the results on the validation set. Since the author-released model of Cylinder3D~\cite{zhu2021cylindrical} utilizes SemanticKITTI for pre-training, here, we report its result based on training the model from scratch for a fair comparison. 
\begin{minipage}[c]{.49\textwidth}
    \centering
        \scriptsize
  \centering
    \begin{tabular}{c|c|c}
    \toprule
    \scriptsize{Method} & \multicolumn{1}{p{2em}|}{\scriptsize{Anno.}} & \scriptsize{mIOU(\%)} \\
    \midrule
    \scriptsize{(AF)2-S3Net~\cite{cheng20212}} & \multirow{5}[2]{*}{\scriptsize{100\%}} & \scriptsize{62.2} \\
    \scriptsize{SPVNAS~\cite{tang2020searching}} &       & \scriptsize{74.8} \\
    \scriptsize{\textbf{Cylinder3D}~\cite{zhu2021cylindrical}} &       & \scriptsize{75.4} \\
    \scriptsize{AMVNet~\cite{liong2020amvnet}} &       & \scriptsize{77.2} \\
    \scriptsize{RPVNet~\cite{xu2021rpvnet}} &       & \scriptsize{77.6} \\
    \midrule
    \scriptsize{ContrastiveSC~\cite{hou2021exploring}} & \scriptsize{0.2\%} & \scriptsize{63.5}  \\
    \scriptsize{LESS (Ours)} & \scriptsize{0.2\%} & \scriptsize{\textbf{73.5}} \\
    \scriptsize{ContrastiveSC~\cite{hou2021exploring}} & \scriptsize{0.9\%} & \scriptsize{65.5}  \\
    \scriptsize{LESS (Ours)} & \scriptsize{0.9\%} & \scriptsize{\textbf{74.8}} \\
    \bottomrule
    \end{tabular}%
    \captionof{table}{\textbf{Comparison on nuScenes validation set.} Cylinder3D~\cite{zhu2021cylindrical} is our fully supervised counterpart.}
  \label{tab:nuscenes}%

\end{minipage}%
\begin{minipage}[c]{.02\textwidth}
    \quad
\end{minipage}%
\begin{minipage}[r]{0.49\textwidth}
    \centering
        \scriptsize
  \centering
    \begin{tabular}{cccccc}
    \toprule
    \scriptsize{pre-} & \scriptsize{weak}  & \scriptsize{propa.}  & \scriptsize{proto.} & \scriptsize{multi-scan} & \scriptsize{mIoU} \\
    \scriptsize{seg.} &  \scriptsize{labels} &  \scriptsize{labels} &  \scriptsize{learning} &  \scriptsize{distillation} & \scriptsize{(\%)} \\
    \midrule
    {\color{gray}\xmark}     & {\color{gray}\xmark}     & {\color{gray}\xmark}     & {\color{gray}\xmark}     & {\color{gray}\xmark}     & 48.1 \\
    \cmark     & {\color{gray}\xmark}     & {\color{gray}\xmark}     & {\color{gray}\xmark}     & {\color{gray}\xmark}     & 59.3 \\
    \cmark     & \cmark     &  {\color{gray}\xmark}     & {\color{gray}\xmark}     & {\color{gray}\xmark}     & 61.6 \\
    \cmark     & {\color{gray}\xmark}     & \cmark     & {\color{gray}\xmark}     & {\color{gray}\xmark}     &  62.2\\
    \cmark     & \cmark     & \cmark     & {\color{gray}\xmark}     & {\color{gray}\xmark}     &  63.5\\
    \cmark     & \cmark     & \cmark     & \cmark     & {\color{gray}\xmark}     & 64.9 \\
    \cmark     & \cmark     & \cmark     & \cmark     & \cmark     & 66.0 \\
    \bottomrule
    \end{tabular}%
  \captionof{table}{\textbf{Ablation study on the SemanticKITTI validation set.} All variants use 0.1\% sparse labels.}
   \label{tab:ablation}%

\end{minipage}
\vspace{1em}

\noindent For other fully-supervised methods~\cite{cheng20212,tang2020searching,liong2020amvnet,xu2021rpvnet}, the results are either obtained from the literature or correspondences with the authors. Since no prior label-efficient work is tested on the nuScenes~\cite{caesar2020nuscenes} dataset, we adapt the source code published by the authors to train ContrastiveSceneContext~\cite{hou2021exploring} from scratch.

\begin{wrapfigure}{r}{0.38\textwidth}
  \centering
  \vspace{-2.5em}
   \includegraphics[width=\linewidth]{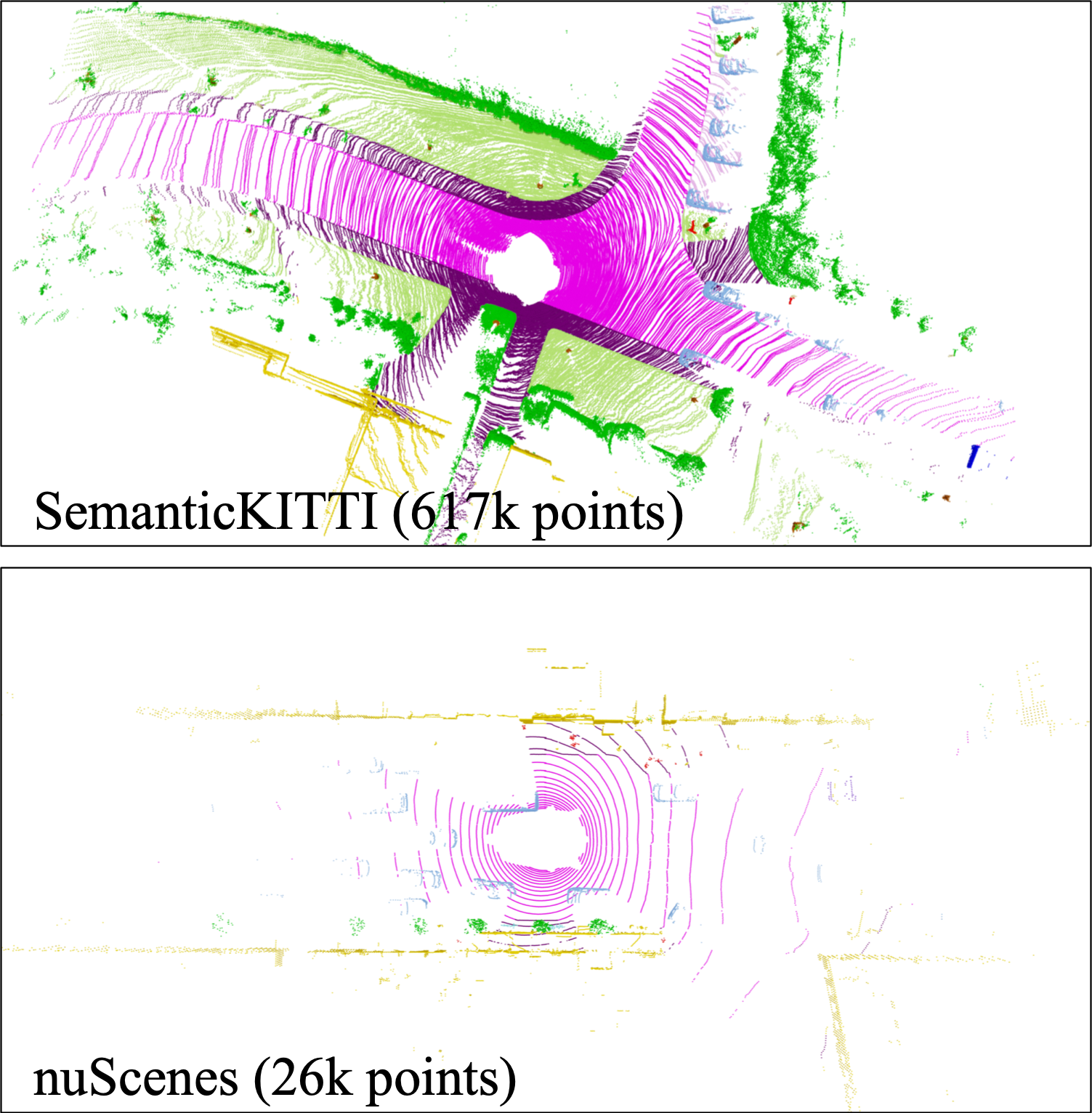}
   \vspace{-2.5em}
   \label{fig:pilot_study}
\end{wrapfigure}

We want to point out that points in the nuScenes dataset are much sparser than those in SemanticKITTI. In nuScenes, only 2 scans per second are labeled, while in SemanticKITTI, 10 scans per second are labeled. Due to the difference of sensors (32-beam vs.\ 64-beam), the number of points per scan in nuScenes is also much smaller (26k vs.\ 120k). See the right inset for the comparison of two datasets (fused points for 0.5 seconds). Considering the sparsity of the original ground truth labels, here we report the 0.2\% and 0.9\% annotation settings. 

\cref{tab:nuscenes} shows the results, where our proposed method outperforms ContrastiveSceneContext~\cite{hou2021exploring} by a large margin. With only 0.2\% sparse labels, our result is also highly competitive with the fully-supervised counterpart~\cite{zhu2021cylindrical}.

\vspace{-1em}
\subsection{Ablation study}
\vspace{-0.5em}
\label{sec:exp-ablation}

\cref{tab:ablation} shows the ablation study of each component. The first row is the result of training with 0.1\% random point labels. By incorporating the pre-segmentation, we spend the limited annotation budget on more underrepresented instances, thereby significantly increasing mIoU from 48.1\% to 59.3\%. Derived from the component proposals, weak labels and propagated labels complement the human-annotated sparse labels and provide dense supervision. Compared to multi-category weak labels, propagated labels provide more accurate supervision and thus lead to a slightly higher gain. Both contrastive prototype learning and multi-scan distillation further boost the performance and finally close the gap between LESS and the fully-supervised counterpart in terms of mIoU.

\vspace{-1em}
\subsection{Analysis of pre-segmentation \& labeling}
\vspace{-0.5em}
\label{sec:exp-pre-seg}

\begin{table}[t]
  \scriptsize
  \centering
    \begin{tabular}{c|c|c}
    \toprule
    Statistics & SemanticKITTI & nuScenes \\
    \midrule
    one-category components & 68.6\% & 80.6\% \\
    two-category components & 23.8\% & 14.9\% \\
    components with more than two categories & 7.6\% & 4.5\% \\
    average number of categories per component & 1.40  & 1.25 \\ \hline
    coverage of sparse labels & 0.1\% & 0.2\% \\
    coverage of propagated labels  & 42.0\% & 53.6\% \\
    coverage of weak labels & 95.5\% & 99.0\% \\
    \bottomrule
    \end{tabular}%
  \caption{\textbf{Statistics of the pre-segmentation and labeling.} Only sparse labels are directly annotated by humans.}
  \label{tab:pre-segmentation}%
\end{table}%

\begin{table}[t]
  \centering
  \scriptsize
    \begin{tabular}{c|c|c|c|c}
    \toprule
      & pre-seg- & mIoU & \#labels for & IoU (\%) of \\
     annotation policy & mentation &    (\%)   &  motorcycle &  motorcycle \\
    \midrule
    randomly sample points &  {\color{gray}\xmark}      & 48.1 &    943   & 22.2 \\
    randomly sample scans &  {\color{gray}\xmark}      & 35.6 &    548   & 0.0 \\
    active labeling~\cite{hou2021exploring} & {\color{gray}\xmark}       &   54.2    &   456    & 36.7 \\
    uniform grid partition &    \cmark   &  61.4     &  1024 (76k)    & 61.3 \\
    geometric partition~\cite{landrieu2018large,liu2021one} &  \cmark & 61.9          &   1190 (294k)    & 64.0 \\
    LESS (Ours) &   \cmark    & 64.9 &    1146 (933k)   & 72.3 \\
    \bottomrule
    \end{tabular}%
\caption{\textbf{Comparison of various annotation policies on SemanticKITTI.} All methods utilize 0.1\% annotations and the same backbone network~\cite{zhu2021cylindrical}. The fourth column indicates the number of sparse labels (and propagated labels) for an underrepresented category (\ie, motorcycle). Multi-scan distillation is not utilized here. The IoU results are calculated on the validation set.}
  \label{tab:annotation}
\end{table}%

By leveraging the unique geometric structure and a careful design, our pre-segmentation works well for outdoor LiDAR point clouds. \cref{tab:pre-segmentation} summarizes some statistics of the pre-segmentation and labeling results. For both datasets, only less than $10\%$ of the components contain more than two categories, which validates that our pre-segmentation generates high-purity components. The high ``coverage of propagated labels'' indicates that we thus deduce a good amount of ``free'' supervision from the pure components. The low ``coverage of sparse labels'' shows that annotators indeed only need to label a tiny portion of points, thus reducing human effort. The ``coverage of weak labels'' confirms that the proposed components can faithfully cover most points. Furthermore, the consistent results across two distinct datasets verify that our method generalizes well in practice.

\cref{tab:annotation} shows the comparison of different annotation policies (\ie, how to use the labeling budget). The first two baselines are introduced in~\cref{sec:pilot_study}, ``active labeling'' utilizes the features from contrastive pre-training to actively select points~\cite{hou2021exploring}, ``uniform grid partition'' uniformly divides the fused point clouds into a grid according to the $xy$ coordinates and treats each cell as a component, ``geometric partition'' extracts handcrafted geometric features and solves a minimal partition problem~\cite{landrieu2018large,zhu2021cylindrical}. All of them are trained with the same backbone Cylinder3D~\cite{zhu2021cylindrical}. The first three methods employ no pre-segmentation and are trained with $\mathcal{L}_{\text{sparse}}$ only. The other approaches utilize our labeling policy (\ie, one label per class
for each component) and are trained with additional $\mathcal{L}_{\text{propagated}}$, $\mathcal{L}_{\text{weak}}$, and $\mathcal{L}_{\text{proto}}$. As a result, their performances are much higher than the first three methods. We also report the number of labels and the IoU for an underrepresented category. We see that our policy leads to more useful supervisions and higher IoUs for underrepresented categories. 

\vspace{-1em}
\subsection{Analysis of multi-scan distillation}
\vspace{-0.5em}
\label{sec:exp-multi-scan}

\begin{figure*}[t]
  \centering
   \includegraphics[width=\linewidth]{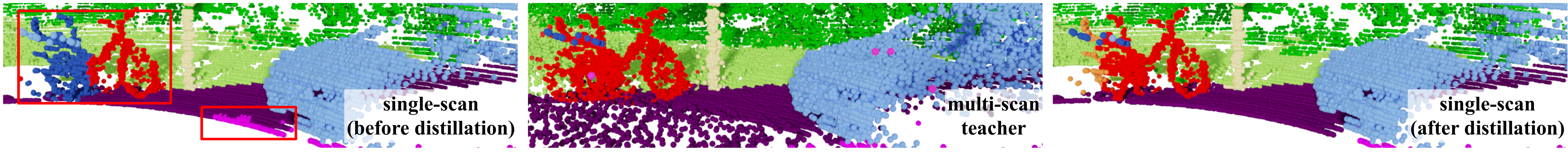}
   \caption{\textbf{Improving segmentation with multi-scan distillation.} 
   The multi-scan teacher leverages the richer semantics via temporal fusion to accurately segment the bicycle and ground, which provides high-quality supervision to enhance the single-scan model. }
   \label{fig:distillation}
\end{figure*}

\begin{table}[t]
  \scriptsize
  \centering
    \begin{tabular}{c|c|c|c|c|c}
    \toprule
    \multicolumn{2}{c|}{single-scan (before)  } & \multicolumn{2}{c|}{multi-scan teacher} & \multicolumn{2}{c}{single-scan (after)} \\
    \hline
    mIoU  & bicycle & mIoU  & bicycle & mIoU  & bicycle \\
    \midrule
    64.9\% &  45.6\%     &  66.8\%     &    51.5\%   & 66.0\% & 49.9\% \\
    \bottomrule
    \end{tabular}%
    \caption{\textbf{Results of the multi-scan distillation on the SemanticKITTI validation set.} 0.1\% annotations are used.}
    \vspace{-2em}
  \label{tab:multi-scan}%
\end{table}%

\cref{tab:multi-scan} and \cref{fig:distillation} show the results of multi-scan distillation. The teacher model exploits the densified point clouds via temporal fusion and thus performs better than the single-scan model (even compared to the fully supervised single-scan model). Through knowledge distillation from the teacher model, the student model improves a lot in the underrepresented classes and completely matches the fully supervised model in mIoU.

\vspace{-0.7em}
\section{Conclusion and future work}
\vspace{-0.7em}

We study label-efficient LiDAR point cloud semantic segmentation and propose a pipeline that co-designs the labeling and the model learning and can work with most 3D segmentation backbones. We show that our method can utilize bare minimum human annotations to achieve highly competitive performance.

We have shown LESS is an effective approach for bootstrapping labeling and learning from scratch. In addition, LESS is also highly compatible for efficiently improving a performant model. 
With the predictions of an existing model, the proposed pipeline can be used for annotators to pick and label component proposals of high-values, such as underrepresented classes, long-tail instances, classes with most failures, \etc. We leave this for future exploration.

\appendix

\renewcommand{\thesection}{S}
\renewcommand{\thetable}{S\arabic{table}}
\renewcommand{\thefigure}{S\arabic{figure}}
\renewcommand{\theequation}{S.\arabic{equation}}

\newpage

In this supplementary material, we first present the implementation and training details of our proposed method and baseline methods (\cref{sec:sup-implementation}). We then show the visual examples of our pre-segmentation results (\cref{sec:sup-pre-seg}), the full results on the nuScenes dataset (\cref{sec:sup-nusc}), and the multi-scan distillation results on the SemanticKITTI dataset (\cref{sec:sup-multi-scan}). Finally, we analyze the generated label distribution (\cref{sec:sup-label}) and the robustness to label noise (\cref{sec:label-noise}). 

\subsection{Implementation \& training details}
\label{sec:sup-implementation}

\subsubsection{Pre-segmentation \& labeling}

While some prior works require perfect pre-segmentation results, our proposed labeling and training pipeline (using weak and propagated labels) allows imperfect component proposals (e.g., a component with multiple categories or an object instance divided into multiple components), which greatly mitigates the impact of pre-segmentation quality on final performance. Our pre-segmentation heuristic only includes two key steps: ground removal and connected component construction. Compared to other complex heuristics, it has fewer hyperparameters. Also, thanks to the good property of outdoor point clouds (i.e., objects are well-separated), we find that, in our experiments, the hyper-parameters are intuitive and easy to select without much effort. 

For example, during the ground removal, we find that the cell size and the RANSAC threshold are robust across datasets, and we set them to be $5m\times 5m$ and $0.2m$ for both datasets. When building connected components, the parameter $d$ should accommodate the LiDAR sensor (the sparser the points, the larger the $d$). We set $d$ to $0.01$ and $0.02$ for SemanticKITTI~\cite{behley2019semantickitti} and nuScenes~\cite{caesar2020nuscenes} datasets, respectively. In our experiments, choosing hyper-parameters with visual inspection is convenient and sufficient to achieve satisfactory results.

\begin{wrapfigure}{r}{0.38\textwidth}
  \centering
  \vspace{-2em}
   \includegraphics[width=\linewidth]{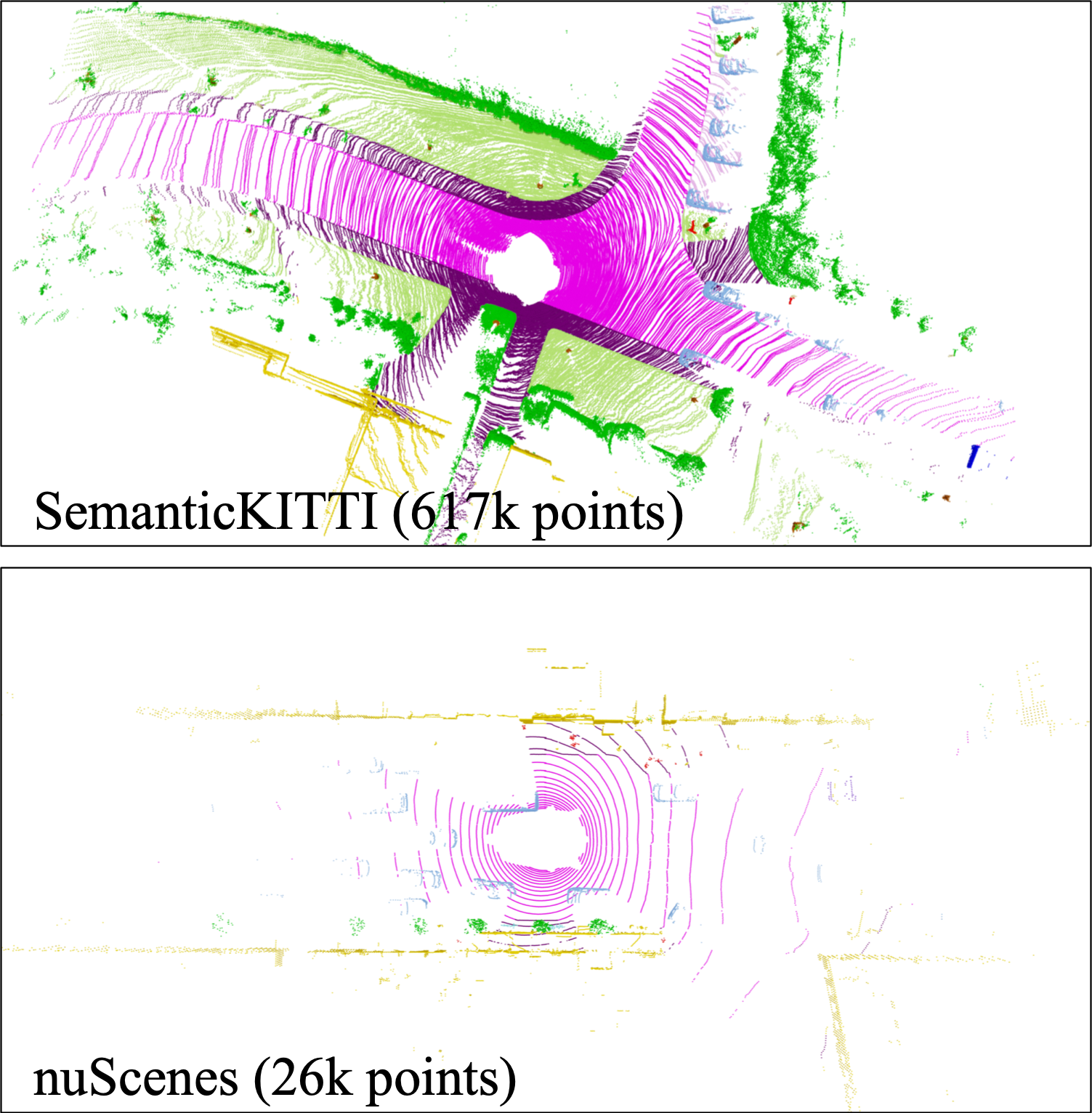}
   \vspace{-2.5em}
   \label{fig:pilot_study}
\end{wrapfigure}
For the SemanticKITTI~\cite{behley2019semantickitti} dataset, we fuse every 5 adjacent scans for the 0.1\% setting and every 100 adjacent scans for the 0.01\% setting. Fusing more adjacent scans will improve labeling efficiency, but may sacrifice pre-segmentation quality as points may become blurry, especially for dynamic objects. After constructing connected components, oversized components are subdivided along the $xy$ axes to ensure each component is within a fixed size (i.e., $2m\times2m$ for non-ground components). We also ignore small components with no more than 100 points. For each component of size $s$, we randomly label 1 point for each category whose number of points is more than $0.05s$. The motivation here is to prevent those noisy and ambiguous points within each component from decreasing the component purity. In real applications, human labelers may also miss or ignore those noisy categories to accelerate the annotation.

For the nuScenes~\cite{caesar2020nuscenes} dataset, we share the same hyperparameters as SemanticKITTI, except for the following. We fuse every 40 adjacent scans, and ignore small components with no more than 10 points. For each component proposal of size $s$, we randomly label 1 (or 4) point(s) for each category whose number of points is more than $0.01s$, corresponding to the  0.2\% (0.9\%) settings. These subtle differences are mainly due to the points in the nuScenes~\cite{caesar2020nuscenes} dataset are much sparser (e.g., the right inset shows the fused points for 0.5 seconds), and we fuse more points and annotate more labels to compensate for the point sparsity.

\subsubsection{Network training}

As for contrastive prototype learning, the momentum parameter $m$ is empirically set to 0.99, temperature parameter $\tau$ is set to 0.1. In multi-scan distillation, we fuse the scans at time $\{t+0.5i ; i \in[-2,2]\}$ for SemanticKITTI, and $\{t+0.5i ; i \in[-3,3]\}$ for nuScenes. We tried multiple sets of parameters (different numbers of scans and intervals). They do lead to some differences ($\sim$3\% mIOU), and we choose the best empirically. We keep all points for scan $i = 0$, and use voxel downsampling to sub-sample $120k$ points from other scans. The temperature $T$ is set to 4. 

We sum up all loss terms with equal weights and train the models on 4 NVIDIA A100 GPUs. For SemanticKITTI, the batch size is 12 and 8 for the single-scan and the multi-scan model, respectively. For nuScenes, the batch size is 16 and 12 for the single-scan and the multi-scan model, respectively.
We utilize the Adam optimizer, and the learning rate is initially set to 1e-3 and then decayed to 1e-4 after convergence. During distillation, the learning rate is set to 1e-4. Other training parameters are the same as Cylinder3D~\cite{zhu2021cylindrical}.

\subsubsection{Baseline Methods}

\begin{figure}[t]
  \centering
   \includegraphics[width=\linewidth]{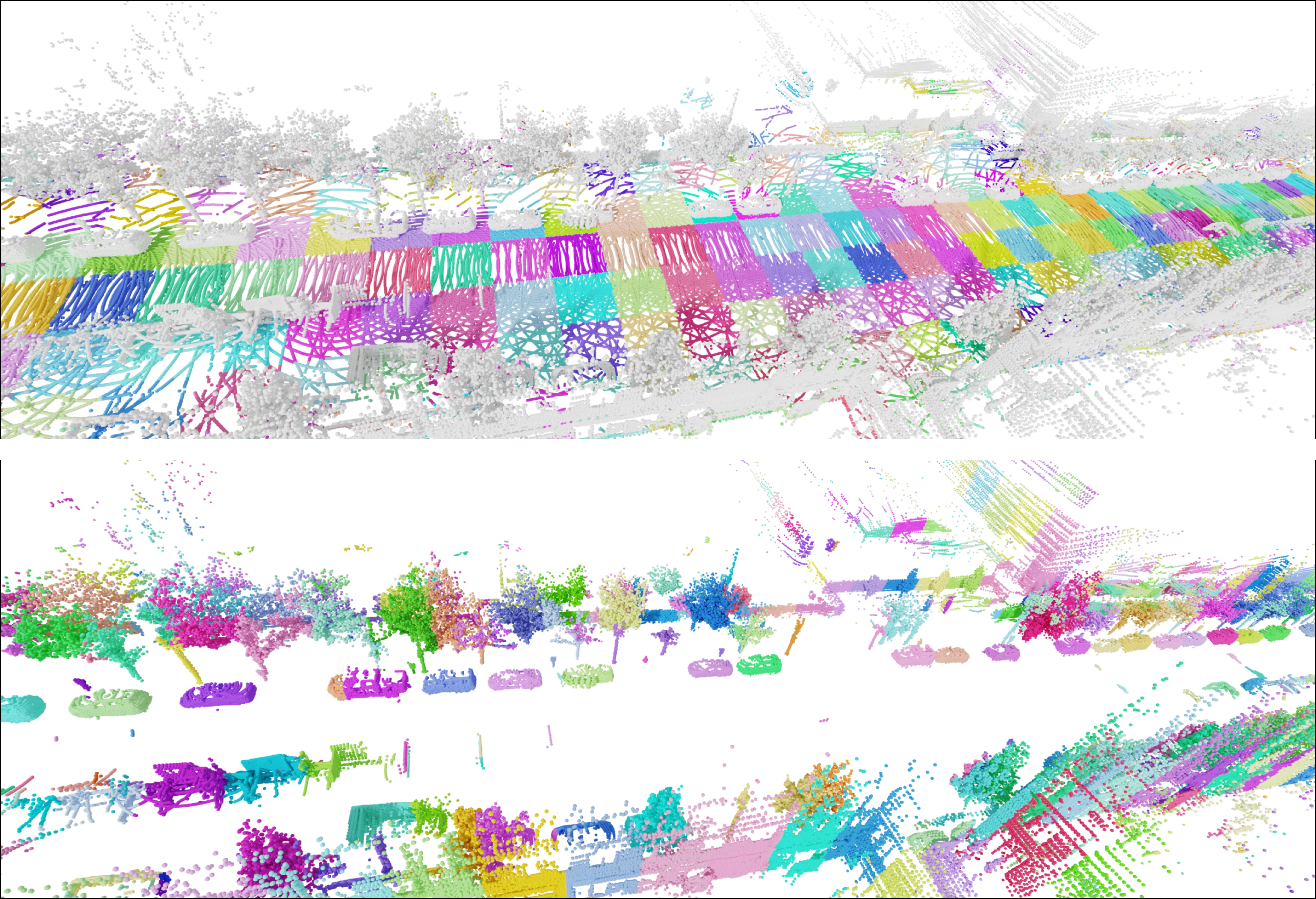}
   \caption{\textbf{Examples of the pre-segmentation results.} First row: detected ground points of each cell. Non-ground points are colored in gray. Each other color indicates a proposed ground component. Second row: connected components of the non-ground points. Each color indicates a connected component. The example is from the nuScenes dataset, where 40 scans are fused.}
   \label{fig:pre-segmentation}
\end{figure}

\setlength\tabcolsep{0.8pt}
\begin{table*}[t]
  \centering
  \scriptsize
    \begin{tabular}{c|c|c|cccccccccccccccc}
    \toprule
    Method & Anno. & mIoU  & \begin{sideways}barrier\end{sideways} & \begin{sideways}bicycle\end{sideways} & \begin{sideways}bus\end{sideways} & \begin{sideways}car\end{sideways} & \begin{sideways}construction-vehicle\end{sideways} & \begin{sideways}motorcycle\end{sideways} & \begin{sideways}pedestrian\end{sideways} & \begin{sideways}traffic-cone\end{sideways} & \begin{sideways}trailer\end{sideways} & \begin{sideways}truck\end{sideways} & \begin{sideways}drivable-surface\end{sideways} & \begin{sideways}other-flat\end{sideways} & \begin{sideways}sidewalk\end{sideways} & \begin{sideways}terrian\end{sideways} & \begin{sideways}manmade\end{sideways} & \begin{sideways}vegetation\end{sideways} \\
    \midrule
    (AF)2-S3Net~\cite{cheng20212} & \multirow{8}[2]{*}{100\%} & 62.2  & 60.3  & 12.6  & 82.3  & 80.0  & 20.1  & 62.0  & 59.0  & 49.0  & 42.2  & 67.4  & 94.2  & 68.0  & 64.1  & 68.6  & 82.9  & 82.4 \\
    RangeNet++~\cite{milioto2019rangenet++} &       & 65.5  & 66.0  & 21.3  & 77.2  & 80.9  & 30.2  & 66.8  & 69.6  & 52.1  & 54.2  & 72.3  & 94.1  & 66.6  & 63.5  & 70.1  & 83.1  & 79.8 \\
    PolarNet~\cite{zhang2020polarnet} &       & 71.0  & 74.7  & 28.2  & 85.3  & 90.9  & 35.1  & 77.5  & 71.3  & 58.8  & 57.4  & 76.1  & 96.5  & 71.1  & 74.7  & 74.0  & 87.3  & 85.7 \\
    SPVNAS~\cite{tang2020searching} &       & 74.8  & 74.9  & 39.9  & 91.1  & 86.4  & 45.8  & 83.7  & 72.1  & 64.3  & 62.5  & 83.3  & 96.2  & 72.7  & 73.6  & 74.1  & 88.3  & 87.4 \\
    \textbf{Cylinder3D~\cite{zhu2021cylindrical}} &       & 75.4  & 75.3  & 41.7  & 91.6  & 86.1  & 52.9  & 79.3  & 79.2  & 66.1  & 61.5  & 81.7  & 96.4  & 72.3  & 73.8  & 73.5  & 88.1  & 86.5 \\
    AMVNt~\cite{liong2020amvnet} &       & 77.0  & 77.7  & 43.8  & 91.7  & 93.0  & 51.1  & 80.3  & 78.8  & 65.7  & 69.6  & 83.5  & 96.9  & 71.4  & 75.1  & 75.3  & 90.1  & 88.3 \\
    RPVNet~\cite{xu2021rpvnet} &       & 77.6  & 78.2  & 43.4  & 92.7  & 93.2  & 49.0  & 85.7  & 80.5  & 66.0  & 66.9  & 84.0  & 96.9  & 73.5  & 75.9  & 76.0  & 90.6  & 88.9 \\
    \midrule
    ContrastiveSC~\cite{hou2021exploring} & 0.2\% & 63.5  & 65.6  & 0.0   & 82.7  & 87.3  & 42.8  & 46.3  & 57.1  & 32.2  & 59.0  & 76.4  & 94.2  & 62.5  & 65.9  & 68.8  & 87.8  & 86.8 \\
    LESS (Ours) & 0.2\% & \textbf{73.5} & 73.7  & 38.3  & 92.0  & 89.7  & 46.9  & 75.6  & 70.9  & 58.4  & 64.8  & 83.0  & 95.6  & 67.6  & 70.9  & 71.8  & 89.2  & 87.3 \\
    ContrastiveSC~\cite{hou2021exploring} & 0.9\% & 64.5  & 64.0  & 12.7  & 80.7  & 87.6  & 41.1  & 55.8  & 61.6  & 37.5  & 59.1  & 75.2  & 94.2  & 65.6  & 67.0  & 70.1  & 88.0  & 87.2 \\
    LESS (Ours) & 0.9\% & \textbf{74.8} & 75.0  & 42.3  & 91.9  & 89.9  & 51.0  & 80.0  & 72.6  & 60.1  & 64.9  & 83.6  & 95.7  & 67.5  & 71.7  & 73.1  & 89.5  & 87.6 \\
    \bottomrule
    \end{tabular}%
  \caption{\textbf{Comparison of different methods on the nuScenes validation set.} Cylinder3D~\cite{zhu2021cylindrical} is our fully supervised counterpart.}
  \label{tab:nuscene_full}%
\end{table*}%

We adopt the author released code to train OneThingOneClick~\cite{liu2021one} and ContrastiveSceneContext~\cite{hou2021exploring} on SemanticKITTI and nuScenes. For other methods, the results are either obtained from the literature or correspondences with the authors. 

For \textbf{ContrastiveSceneContext}~\cite{hou2021exploring}, we first compute the overlapping ratio between every pair of scans within each sequence, where the voxel size is set to $0.3m$. We then use pairs of scans whose overlapping ratio is no less than $30\%$ for contrastive pre-training. During pre-training, we train the model with a voxel size of $0.15m$ for 100k iterations. The batch size is 12 and 20 for SemanticKITTI and nuScenes, respectively. We then follow the provided pipeline to infer the point features and select points for labeling. After that, we train the segmentation network with the pre-trained weights for 30k iterations. The voxel size is set to $0.1m$, and the batch size is set to 18 and 36 SemanticKITTI and nuScenes, respectively. We disable the elastic distortion and the color-related data augmentation.

For \textbf{OneThingOneClick}~\cite{liu2021one}, we first apply the geometrical partition described in~\cite{landrieu2018large} to generate the super-voxels, where only the point coordinates are used as input. We then randomly label a subset of super-voxels for a given annotation budget. We follow the authors' guidance to train the modules for three iterations. In each iteration, we train the 3D-U-Net for 32 epochs (51k iterations) and the RelationNet for 64 epochs (102k iterations). During training, the voxel size is set to $0.1m$, and the batch size is set to $12$. We disable the elastic distortion for the data augmentation.  

\subsection{Visual results of pre-segmentation}
\label{sec:sup-pre-seg}
\cref{fig:pre-segmentation} shows the examples of our pre-segmentation results.

\setlength\tabcolsep{2pt}
\begin{table*}[t]
  \centering
  \scriptsize
    \begin{tabular}{c|rrrrrrrrrrrrrrrrrrr} 
    \toprule
          & \multicolumn{1}{c}{\begin{sideways}car\end{sideways}} & \multicolumn{1}{c}{\begin{sideways}bicycle\end{sideways}} & \multicolumn{1}{c}{\begin{sideways}motorcycle\end{sideways}} & \multicolumn{1}{c}{\begin{sideways}truck\end{sideways}} & \multicolumn{1}{c}{\begin{sideways}other-vehicle\end{sideways}} & \multicolumn{1}{c}{\begin{sideways}person\end{sideways}} & \multicolumn{1}{c}{\begin{sideways}bicyclist\end{sideways}} & \multicolumn{1}{c}{\begin{sideways}motorcyclist\end{sideways}} & \multicolumn{1}{c}{\begin{sideways}road\end{sideways}} & \multicolumn{1}{c}{\begin{sideways}parking\end{sideways}} & \multicolumn{1}{c}{\begin{sideways}sidewalk\end{sideways}} & \multicolumn{1}{c}{\begin{sideways}other-ground\end{sideways}} & \multicolumn{1}{c}{\begin{sideways}building\end{sideways}} & \multicolumn{1}{c}{\begin{sideways}fence\end{sideways}} & \multicolumn{1}{c}{\begin{sideways}vegetation\end{sideways}} & \multicolumn{1}{c}{\begin{sideways}trunk\end{sideways}} & \multicolumn{1}{c}{\begin{sideways}terrain\end{sideways}} & \multicolumn{1}{c}{\begin{sideways}pole\end{sideways}} & \multicolumn{1}{c}{\begin{sideways}traffic sign\end{sideways}}\\ 
    \midrule
    sparse  ($\times$0.1\%) & 0.8  & \textbf{2.7}  & 0.9  & 0.6  & 0.8  & \textbf{1.8}  & \textbf{1.8}  & \textbf{2.7}  & 0.4  & 0.7  & 0.6  & 1.4  & 0.8  & 1.0  & 1.0  & \textbf{2.0}  & 1.0  & \textbf{3.1}  & \textbf{4.1} \\ 
    propagated  (\%) & 79  & 12  & 75  & 77  & 75  & 52  & 64  & 48  & 16  & 6   & 9   & 17  & 77  & 25  & 55  & 29  & 32  & 28  & 9  \\ 
    \bottomrule
    \end{tabular}%
\caption{\textbf{The coverage of sparse labels and propagated labels for the SemanticKITTI dataset.} The numbers are the ratios between the number of sparse labels (and propagated labels) and the number of points within each category.}
  \label{tab:kitti_labels}%
\end{table*}%

\setlength\tabcolsep{3pt}
\begin{table*}[t]
  \centering
  \scriptsize
    \begin{tabular}{c|rrrrrrrrrrrrrrrr} 
    \toprule
          & \multicolumn{1}{c}{\begin{sideways}barrier\end{sideways}} & \multicolumn{1}{c}{\begin{sideways}bicycle\end{sideways}} & \multicolumn{1}{c}{\begin{sideways}bus\end{sideways}} & \multicolumn{1}{c}{\begin{sideways}car\end{sideways}} & \multicolumn{1}{c}{\begin{sideways}construction-vehicle\end{sideways}} & \multicolumn{1}{c}{\begin{sideways}motorcycle\end{sideways}} & \multicolumn{1}{c}{\begin{sideways}pedestrian\end{sideways}} & \multicolumn{1}{c}{\begin{sideways}traffic-cone\end{sideways}} & \multicolumn{1}{c}{\begin{sideways}trailer\end{sideways}} & \multicolumn{1}{c}{\begin{sideways}truck\end{sideways}} & \multicolumn{1}{c}{\begin{sideways}drivable-surface\end{sideways}} & \multicolumn{1}{c}{\begin{sideways}other-flat\end{sideways}} & \multicolumn{1}{c}{\begin{sideways}sidewalk\end{sideways}} & \multicolumn{1}{c}{\begin{sideways}terrian\end{sideways}} & \multicolumn{1}{c}{\begin{sideways}manmade\end{sideways}} & \multicolumn{1}{c}{\begin{sideways}vegetation\end{sideways}} \\ 
    \midrule
    sparse  ($\times$0.1\%) & 2.4  & \textbf{20.9} & 4.0  & 4.6  & 4.8  & \textbf{8.0} & \textbf{19.9} & \textbf{12.2} & 3.4  & 3.4  & 0.6  & 1.7  & 1.9  & 3.1  & 4.8  & 7.9  \\ 
    propagated (\%) & 16  & 16  & 53  & 52  & 54  & 46  & 29  & 20  & 49  & 59  & 32  & 2   & 2   & 11  & 62  & 55  \\ 
    \bottomrule
    \end{tabular}%
  \caption{\textbf{The coverage of sparse labels and propagated labels for the nuScenes dataset.} The numbers are the ratios between the number of sparse labels (and propagated labels) and the number of points within each category.}
  \label{tab:nuscene_label}%
\end{table*}%

\setlength\tabcolsep{2.5pt}
\begin{table*}[t]
  \centering
    \scriptsize
    \begin{tabular}{c|c|ccccccccccccccccccc}
    \toprule
    Method & mIOU  & \begin{sideways}car\end{sideways} & \begin{sideways}\textbf{bicycle}\end{sideways} & \begin{sideways}motorcycle\end{sideways} & \begin{sideways}\textbf{truck}\end{sideways} & \begin{sideways}other-vehicle\end{sideways} & \begin{sideways}\textbf{person}\end{sideways} & \begin{sideways}\textbf{bicyclist}\end{sideways} & \begin{sideways}motorcyclist\end{sideways} & \begin{sideways}road\end{sideways} & \begin{sideways}parking\end{sideways} & \begin{sideways}sidewalk\end{sideways} & \begin{sideways}other-ground\end{sideways} & \begin{sideways}building\end{sideways} & \begin{sideways}fence\end{sideways} & \begin{sideways}vegetation\end{sideways} & \begin{sideways}trunk\end{sideways} & \begin{sideways}terrain\end{sideways} & \begin{sideways}pole\end{sideways} & \begin{sideways}traffic sign\end{sideways} \\
    \midrule
    single-scan (before) & 64.9  & 97  & \textbf{46} & 72  & \textbf{91} & 69  & \textbf{73} & \textbf{88} & 0   & 92  & 39  & 77  & 4   & 90  & 58  & 88  & 66  & 73  & 61  & 52 \\
    multi-scan teacher & 66.8  & 97  & \textbf{52} & 82  & \textbf{94} & 72  & \textbf{78} & \textbf{92} & 0   & 93  & 40  & 79  & 1   & 89  & 54  & 87  & 70  & 72  & 64  & 53 \\
    single-scan (after) & 66.0  & 97  & \textbf{50} & 73  & \textbf{94} & 67  & \textbf{76} & \textbf{92} & 0   & 93  & 40  & 79  & 3   & 91  & 60  & 87  & 68  & 71  & 62  & 51 \\
    \bottomrule
    \end{tabular}%
  \caption{\textbf{Results of the multi-scan distillation on the SemanticKITTI validation set.} 0.1\% annotations are used.}
  \label{tab:multi-scan-full}%
\end{table*}%

\subsection{Full results on nuScenes}

\label{sec:sup-nusc}
\cref{tab:nuscene_full} shows the full results on the nuScenes validation set.

\subsection{Full table of multi-scan distillation}

\label{sec:sup-multi-scan}
\cref{tab:multi-scan-full} shows the full results of the multi-scan distillation. The multi-scan teacher model leverages the richer semantics via temporal fusion and achieves significantly better performances in the underrepresented categories, such as bicycle, person, and bicyclist. Through knowledge distillation from the teacher model, the student model also improves a lot in those categories.

\subsection{Label distribution}

\label{sec:sup-label}

\cref{tab:kitti_labels} and \cref{tab:nuscene_label} summarize the distributions of the generated sparse labels and the propagated labels. By leveraging our proposed pre-segmentation and labeling policy, we put more emphasis on the underrepresented categories. For example, the ratios of sparse labels for bicycle and road are 2.68 vs.\ 0.36 in the SemanticKITTI dataset, and 20.85 vs.\ 0.63 in the nuScenes dataset. As for the propagated labels, we find the distributions are unbalanced. For categories, such as car and building, they are easier to be separated and form pure components, thus having high coverages of propagated labels. However, some categories, such as bicycle, road, sidewalk, and parking, are prone to be connected with other categories, thus having low coverages of propagated labels. The discrepancy between the distributions of the two types of labels confirms that we need to treat them separately instead of simply merging them with a single loss function.

 \subsection{Robustness to label noise}
 \label{sec:label-noise}
 
In the paper, we use point labels from the original datasets to mimic the annotation policy, and no extra noise is added. 

To evaluate the robustness of our method to label noise, we randomly change $3\%$ (or $10\%$) of the sparse point labels to a random category, which alters weak labels and propagated labels accordingly. The resulting mIoU drops $2.1\%$ (or $3.7\%$), which is within a reasonable range and verifies that our method will not be significantly affected by the label noise.

\bibliographystyle{splncs04}
\bibliography{egbib}
\end{document}